\documentclass[runningheads]{llncs}
\usepackage[english]{babel}
\usepackage[utf8]{inputenc}
\usepackage{wrapfig}
\usepackage{adjustbox}
\usepackage{graphicx}
\usepackage{multicol}
\usepackage{algorithmic}
\usepackage{color}
\usepackage{subcaption}
\usepackage[]{algorithm2e}
\usepackage[table]{xcolor}
\usepackage{amssymb}
\usepackage{pifont}
\SetCommentSty{mycommfont}
\SetKwInput{KwInput}{Input}                
\SetKwInput{KwOutput}{Output}              
\usepackage{amsmath}
\usepackage{gensymb}
\usepackage{textcomp}
\usepackage{tikz}

\date{May 2021}
\begin{document}
\title{Tackling COVID-19 Infodemic using Deep Learning}

\author{Prathmesh Pathwar\inst{1}\orcidID{0000-0002-2306-6841} \and
Simran Gill\inst{2}\orcidID{0000-0002-6680-7013}}

\institute{Indian Institute of Information Technology, Allahabad\\
\email {iit2016028@iiita.ac.in, iit2016088@iiita.ac.in}}

\maketitle

\begin{abstract}
Humanity is battling one of the most deleterious virus in modern history, the COVID-19 pandemic, but along with the pandemic there's an infodemic permeating the pupil and society with misinformation which exacerbates the current malady. We try to detect and classify fake news on online media to detect fake information relating to COVID-19 and coronavirus. The dataset contained fake posts, articles and news gathered from fact checking websites like politifact whereas real tweets were taken from verified twitter handles. We incorporated multiple conventional classification techniques like Naive Bayes, KNN, Gradient Boost and Random Forest along with Deep learning approaches, specifically CNN, RNN, DNN and the ensemble model RMDL. We analyzed these approaches with two feature extraction techniques, TF-IDF and GloVe Word Embeddings which would provide deeper insights into the dataset containing COVID-19 info on online media.

\end{abstract}
\section{Introduction}
Misinformation is defined as objectively incorrect information that is not supported by scientific evidence and expert opinion \cite{b1}. Since COVID-19 was declared as a pandemic of global scale in March 2020, there has been speculations and theories about the cause, symptoms, treatment, number of cases, fatality levels, the age of infected, conspiracy theories and much more ersatz information revolving in social media causing panic and contention among people. The problem does not engender just from presence of fake news, but the rapid pace at which it spreads among 3.6 billion social media users in 2020 \cite{b2} with numbers exponentially increasing in future. According to Economic Times, Whatsapp (which is now part of Facebook), limited forwarding viral messages to one person down from five \cite{b6}, when Indian government advised the need to control hoaxes on social media platforms as it eventually poses a threat to society \cite{b3}. In an article by Forbes, other companies followed suite by deploying some mechanism to tackle dissemination of fake content and give priority to reliable, credited sources like WHO and government offices \cite{b5}. Further, studies indicated that misinformation was more frequently tweeted than science based evidence \cite{b7} \cite{b8}, concluding that misinformation can stymie the efforts for preventive measures \cite{b9}, which could exacerbate the damage done by pandemic. 

Globally institutions began to realize that fake news regarding COVID-19 poses a similar if not higher threat that the virus itself, with news like \textit{"Alcohol is a cure for COVID-19"} which led to multiple deaths in Iran \cite{b4} and in recent times hate and attack on Asian communities have increased several fold given to the baseless conspiracies regarding China. According to Business Today \cite{b10}, WHO SEARO released a statement in April 2021 that a video of it \textit{predicting 50,000 deaths} in India is fake, pointing to the fact even after a year of battling the \textit{Infodemic}, much is to be done. 

In this paper, we aim to detect the fake news surrounding COVID-19 and Coronavirus on social media platforms. We analyze the dataset containing a 10,700 fake and real tweets. We analyze different conventional and deep learning models using Term-Frequencey Inverse-Document-Frequency (TFIDF) \cite{b11} and Glove word embeddings \cite{b12} to detect fake COVID-19 information efficiently. This analysis would aid institutions to filter information and tag fake news and misinformation at early stage so that it's pernicious effects can be mitigated. Further, this underscores the importance of proper models required to espouse society and medical institutions to save resources and time on tackling misinformation and concentrate towards the pandemic, acting as an anodyne for the infected. ML researchers can obtain greater insight regarding the deep learning and feature extraction paradigms with respect to COVID-19 pandemic. 

\section{Literature Review}

There is a lot of research done regarding fake news classification and detection on social media and multiple approaches have been suggested to improve the models. We look into some of the works done by researchers worldwide on the topic of text classification, fake news detection and approaches using deep learning .

Hadeer Ahmed, Issa Traore, Sherif Saad \cite{b13} used n-gram analysis and different feature extraction techniques: SVM, LSVM and Logistic Regression on the 2 datasets obtained from kaggle.com, tripadvisor.com and other sources to detect fake news and spam opinions (as both are closely related). It was observed that with increase in number of features the accuracy of classification models increased achieving a best accuracy of 90\%.

Jamal Abdul Nasir, Osama Subhani Khan, Iraklis Varlamis\cite{b14} created a hybrid model of CNN and RNN on FA-KES and ISOT dataset to detect fake news. They compared their results with general techniques like Random forest, Logistic Regression, Naive Bayes, KNN and more and also with standalone DNN and CNN models. Their model achieved an accuracy of 0.60 ± 0.007 on FA-KES and 0.99 ± 0.02 on ISOT. 

Jiangfeng Zenga, Yin Zhangb, Xiao Ma \cite{b15} introduced a novel approach to use semantic correlations across multimodalities using images in FND-SCTI and enhance training the learning of text representation via Hierarchical Attention Mechanism. They explored two more approaches using  multimodal variational autoencoder and  multimodal fusion eigenvector which and tested with  Twitter and Weibo fake news dataset to obtain best accuracy among seven competetive approaches.

Mykhailo Granik, Volodymyr Mesyura \cite{b16} used naive bayes algorithm to detect fake news from a dataset created using facebook posts and news pages. They obtained an accuracy of 74\% which indicates that a simple model can still be used to detect and classify fake news.

Yonghun Jang, Chang-Hyeon Park and Yeong-Seok Seo \cite{b17} proposed a fake news analysis modelling method to get best features from Quote retweet that would positively impact fake news detection. They also employed a method to collect, tweets, posts and user details from twitter and format it so that it can be easily used for fake news analysis. 

Aswini Thota, Priyanka Tilak, Simeratjeet Ahluwalia, Nibhrat Lohia \cite{b18} used Dense Neural Network along with a custom tuned TFIDF and BoW approach to detect classify fake news. There accuracy using Word2Vec word embeddings was comparatively lower to TFIDF and BoW, for which they explored possible reasons but eventually there best accuracy was 94.21\% on FNC-1 dataset.

Álvaro Ibrain Rodríguez, Lara Lloret Iglesias\cite{b19} compared the performance of convolutional based model, LSTM based model and BERT on fake news to attain an accuracy of, after hyper-parameter tuning, 0.91 with LSTM based model, 0.94 with Convolutional based model and 0.98 using BERT.

HT Le, C Cerisara, A Denis \cite{b20} explored the efficiency of deep CNNs in classification domain, they discovvered that the models with short and wide CNNs are most efficient at word level, on the other hand deep models perform better when input text is encoded as a sequence of characters but still are worse on average. They also introduced a new model derived from DeepNet for text inputs. 

\section{Dataset And Experimental Setup}
The analysis and prediction of fake news is conducted on the dataset originally provided by Parth et al. \cite{b21} as part of their research and contains fake news regarding \textit{COVID-19} from various fact-checking websites like Politifact,  NewsChecker ,  Boomlive and from tools like Google fact-check-explorer and IFCN chatbot in the form of posts, tweets, articles. The real news is collected from Twitter using verified twitter handles.
The dataset is divided into three parts, train, validate and test. The train dataset contains total 6420 samples, whereas validation and test each contain a data 2140 samples. The distribution is provided in Table 1.

\begin{table}[!htbp]
\centering 
\begin{tabular}{|c|c|c|c|} 
\hline 
\textbf{Attributes}&\textbf{Real}&\textbf{Fake}&\textbf{Total} \\
\hline 
Train & 3360 & 3060 & 6420\\ 
\hline
Validation  &  1120 & 1020 & 2140 \\
\hline
Test  &  1120 & 1020 & 2140 \\
\hline
Total  &  5600 & 5100 & 10700 \\
\hline 
\end{tabular}
\label{tab:hresult}
\begin{center}
\caption{Dataset Distribution} 
\end{center}
\end{table}

\begin{figure*}[ht]
\centering
    \begin{subfigure}{0.32\textwidth}
      \centering
      \includegraphics[scale = 0.27]{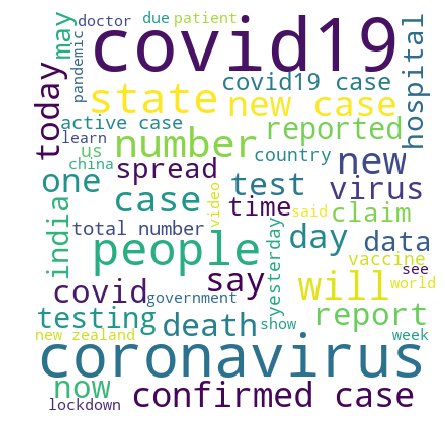}
      \caption{All samples }
      \label{fig:sub-first}
    \end{subfigure}
    \begin{subfigure}{0.32\textwidth}
      \centering
      \includegraphics[scale = 0.27]{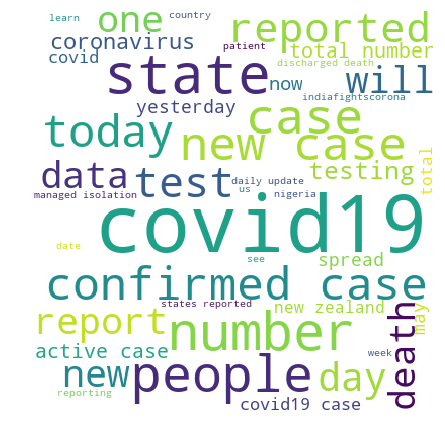}
      \caption{Real samples }
      \label{fig:sub-first}
    \end{subfigure}
    \begin{subfigure}{0.32\textwidth}
      \centering
      \includegraphics[scale = 0.27]{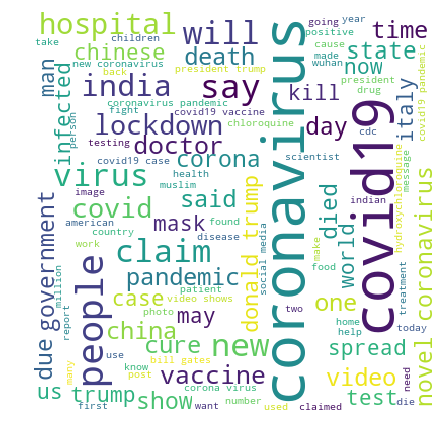}
      \caption{Fake samples}
      \label{fig:sub-first}
    \end{subfigure}
    \caption{Wordclouds for Label Distribution}

\end{figure*}
The dataset has comparable number of Fake and Real samples for both test and train sets, making the training and testing unskewed which would give better insights about the performance of classifiers. 

\section{Methodology}
\subsection{\textbf{\textit{Data Cleaning}}}
Data Cleaning increases the validity and quality of the model by removing incorrect information and refining the data. Raw data might contain null or missing values which might skew the results for our model. 
Following methods were be taken into account while cleaning the data:
\begin{itemize}
    
    \item \textbf{Renaming and removing the columns } : To get name of columns in more readable form, the names are renamed.
    \item \textbf{Renaming the values of a feature}: To form data in consistent and homogeneous format, all data is converted into numerical format.
\end{itemize}
\subsection{\textbf{\textit{Data Pre-processing}}}
The data needs to be preprocessed and refined before before using it to test model through punctuation removal, removing stop words, stemming, lower casing comments, lemmatization and tokenization.
A generic function was used to remove punctuation, hyperlinks and lower-casing the text.\\
Original sample: {\color{blue}\texttt{Our daily update is published. States reported 734k tests 39k new cases and 532 deaths. Current hospitalizations fell below 30k for the first time since June 22. https://t.co/wzSYMe0Sht}
}
\subsection{\textbf{\textit{Removing stop words}}}
Stop words are words that create unnecessary noise when training models. These words may be pronouns, conjunctions, prepositions, hyperlinks or any other words that might be inconsistent with the dataset and hamper the results and training process. We used the nltk's stopwords which contains 179 stopwords specifically for english language.\\  
After removing stopwords: {\color{blue}\texttt{daily update published states reported 734k tests 39k new cases 532 deaths current hospitalizations fell 30k first time since june 22}}
\subsection{\textbf{\textit{Stemming}}}
Stemming essentially is morphing the words into their original form, such as "swimming" is converted to "swim", this removes the multiple variation of words in different formats and boils them down to their meaning for efficient and better learning. We use nltk's Snowball stemmer which is an improved version of Porter stemmer.\\
After stemming: {\color{blue}\texttt{daili updat publish state report test new case death current hospit fell first time sinc june}}
\subsection{\textbf{\textit{Feature Extraction}}}

Feature Extraction is a major step as it adversely affect the performance of the model. Words cannot be used in raw format and hence need to be converted to vectors to enable models to learn. We take two approaches for this, word embeddings and TF-IDF. 

\subsubsection{\textbf{\textit{Word Embeddings}}}
Word Embeddings are pretrained multidimensional vectors which represent relationship among words as vectors. The vectors for "airplane”, “aeroplane”, “plane”, and “aircraft” would be similar as they represent almost the same meaning whereas "airplane" and "book" would have dissimilar vectors. It can be considered that they define relationship between the words, but need large external corpus to obtain the relationships, and are computationally intensive to use as dimensions increase. We use the 50 dimensional twitter GloVe (Global Vectors for Word Representation)\cite{b12}  word vectors which were made by evaluating Wikipedia 2014 and Gigawords 5 translating to 6B tokens, and 400K vocab which were used to obtain embedding with 50, 100, 200 and 300 dimensions. The objective function for it can be defined as
 \begin{equation} \label{eq1}
    \begin{split}
    g(v_i - v_j,\Tilde{v_{j}}) = \frac{P_{ik}}{P_{jk}}
    \end{split}
    \end{equation}
where $v_i$ refers to the word vector of word $i$, and $P_{ik}$ refers to the probability of word $k$ to occur in context of $i$.
\begin{figure}[!htbp]
    \begin{center}
    \includegraphics[scale = 0.16]{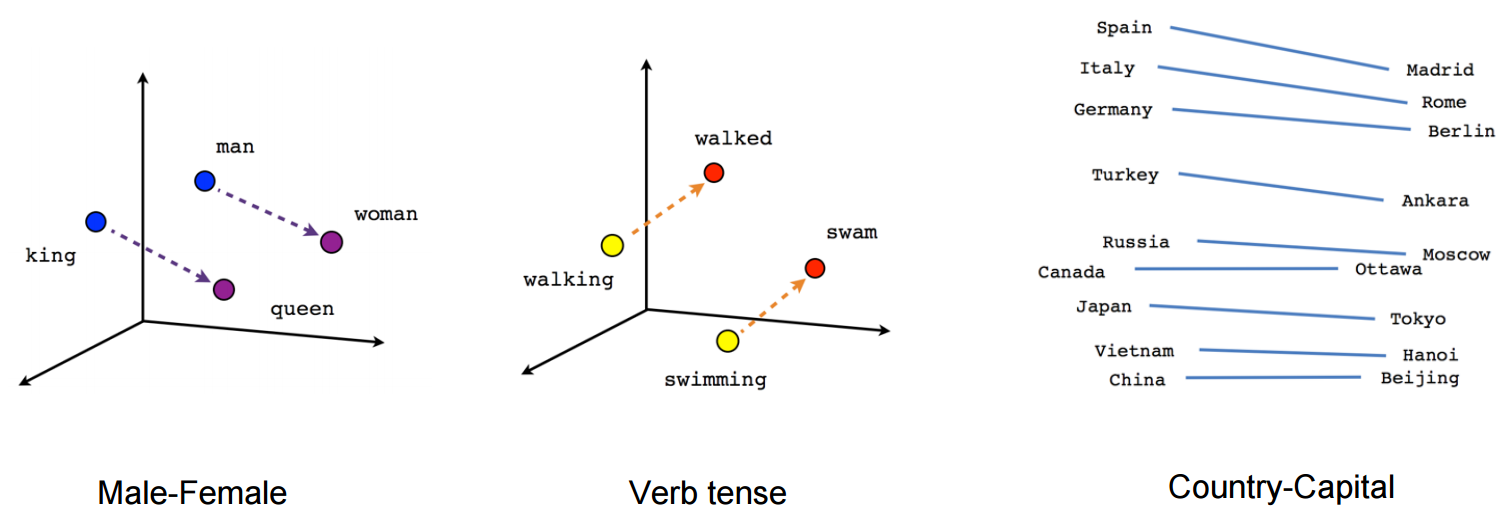}
    \caption{Relationship between vectors}
    \end{center}
\end{figure}

\subsubsection{\textbf{\textit{Term Frequency — Inverse Document Frequency}}}

Term Frequency — Inverse Document Frequency (TF-IDF)\cite{b11} is common tool in NLP for converting a list of text documents to a matrix representation. It is obtained by calculating Term Frequency and Inverse Document Frequency to obtain normalized vectors. The sparse matrix formed measures the times a number appeared in the document and as the count can be large it is also divided by the length of the document as a way of basic normalization. Though obtaining TF-IDF is not computational intensive, it doesn't capture the relationship between words. 
Term Frequency can be defined as the ratio of occurrence of term $t$ in sentence to the number of words in the sentence.
\begin{equation} \label{eq1}
    \begin{split}
    TF(d,t) = \frac{\sum_{t\epsilon d}t}{|d|}
    \end{split}
    \end{equation}
Inverse Document Frequency has a characteristic that it normalizes effect of TF , while raising the significance of rare terms. For example, words like "\textit{the}","\textit{where}" might dominate if only Term Frequency is used but using IDF with TF reduces this impact and gives a normalized output. Here $N$ denotes the number of documents and $f(t)$ denotes the number of documents containing term $t$.
\begin{equation} \label{eq1}
    \begin{split}
    IDF(d,t) = log\left(\frac{(N_d)}{f(t)}\right)
    \end{split}
    \end{equation}

Together we get the TF-IDF matrix.
\begin{equation} \label{eq1}
    \begin{split}
    W(d,t) =TF(d,t) * log\left(\frac{(N_d)}{f(t)}\right)
    \end{split}
    \end{equation}

\subsection{\textbf{\textit{Data Classification using Conventional Models}}}
In this paper, we can broadly categorize the models used. At one part we would use conventional algorithms and on the other hand we will explore deep learning models and networks, to make sure that the dataset is evaluated from multiple perspectives. We used Multinomial Naive Bayes, Gradient Boosting classifier, K Nearest Neighbours and Random Forest classifier, with optimum hyperparameters to conduct the experiment.

\subsection{\textbf{\textit{Deep Learning Models}}}
Since the advent of Perceptron and Neural Network , there has been progress in the field of machine learning with some models been widely used as the basis for complex and highly specific models. Deep Learning models usually require test and training set from which the models can obtain information and "learn" to be later applied to new inputs. We used 4 different deep learning models - CNN, RNN using GRU and LSTM recurrent units, DNN and RMDL (which is an ensemble model of CNN, RNN and DNN) for our text classification objective. 

\section{Experimental Results}
The dataset was pre divided into train, test and validation sets. It contained 6420 training samples, 2140 validation samples and 2140 test samples. The samples contains "tweet" and their respective label "fake" or "real". The labels were changed to 0 for "fake" and 1 for "real". Further the data contained, hyperlinks, emojis and other noise which was cleaned as part of data pre-processing. The testing is done using two feature selection techniques - TFIDF and Glove word embedding. The models used accuracy as the criteria to optimize and minimize loss function. For deep learning models, sparse categorical crossentropy was used as the loss function and Adam optimizer was used.

\begin{itemize}
    \item \textbf{\textit{Sparse Categorical Crossentropy}} : Derived from cross entropy, this is used when the classes are mutually exclusive, so the values for ith label would be present and other would be zero (hence the term "sparse").
        \begin{equation} \label{eq1}
    \begin{split}
    J(w) = -\frac{1}{N}\sum_{i=1}^{N}\left[y_i log(\hat{y_i}) + (1-y_i)log(1-\hat{y_i})\right]
    \end{split}
    \end{equation}
    where $w$ is the model parameters, ${y_i}$ is true label and $\hat{y_i}$ is predicted label.
    
    \item \textbf{\textit{Accuracy}} : It is the calculated as the proximity between the predicted class label and actual class label or the ratio of correctly predicted (True Positive + True Negative) to total number of samples. Accuracy can be calculated as: 
        \begin{equation} \label{eq1}
    \begin{split}
    Accuracy = \frac{True Positive + True Negative}{Total}    \end{split}
    \end{equation}
\end{itemize}
\begin{figure}[!htbp]
\begin{center}
    \includegraphics[scale = 0.33]{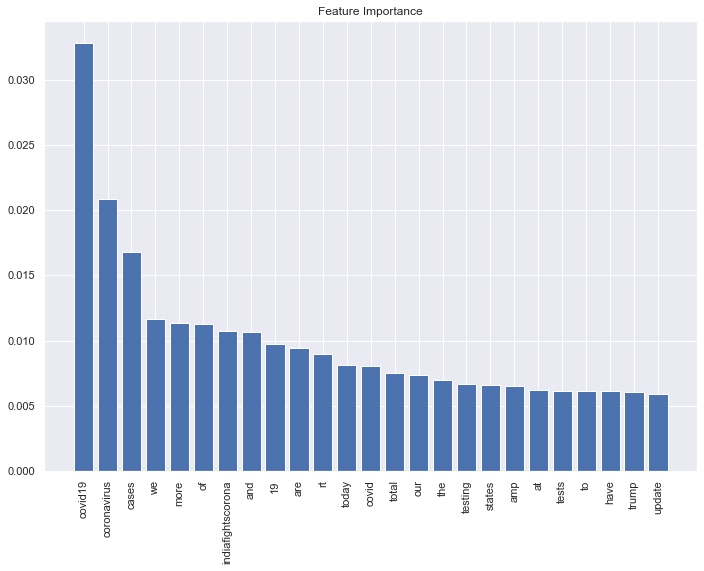}
    \caption{Feature Importance using TFIDF}
    \end{center}
\end{figure}
\subsection{\textbf{\textit{Classification Algorithms for predicting Fake COVID-19 News}}}

\subsubsection{\textbf{\textit{Random Forest Analysis}}}
The two feature extraction techniques were used along-with Random Forest, the number of decision trees (n estimators) used were 64. An analysis was done on the accuracy by using TFIDF and Word Embedding. 

\begin{table}[!htbp]
\begin{center}
    \begin{tabular}{ |c|c|c| } 
    \hline
     \textbf{Metric (in percentage)} & \textbf{TFIDF} & \textbf{Word Embedding} \\
    \hline
     Accuracy & \cellcolor{blue!25}91.30 & 87.05\\
    \hline
     Precision & \cellcolor{blue!25}93.64 & 90.64\\
    \hline
     Recall & \cellcolor{blue!25}89.46 & 83.92\\
    \hline
     F1-Score & \cellcolor{blue!25}91.50 & 87.15\\
    \hline
    \end{tabular}
    \end{center}
        \caption{\label{tab:table-name}Accuracy of Random Forest using TFIDF and Word Embedding}

\end{table}

Thus, the best performance of \textbf{Random Forest} using TFIDF to obtain vectors was 91.30\% and with word embedding it was observed as 87.05\%. Hence TFIDF provided better performance with \textbf{91.30\%} accuracy .

\subsubsection{\textbf{\textit{Multinomial Naive Bayes Analysis}}}
In the case of Multinomial Naive Bayes, the analysis was only done with TFIDF due to the fact that word embedding contained negative vector values which couldn't be used with Naive Bayes as it works on conditional probabilities which are non-negative. 

\begin{table}[!htbp]
\begin{center}
    \begin{tabular}{ |c|c| } 
    \hline
     \textbf{Metric (in percentage)} & \textbf{TFIDF}\\
    \hline
     Accuracy & 90.88\\
    \hline
     Precision & 88.57\\
    \hline
     Recall & 94.82\\
    \hline
     F1-Score & 91.59\\
    \hline
    \end{tabular}
    \end{center}
        \caption{\label{tab:table-name}Accuracy of Multinomial Naive Bayes using TFIDF}

\end{table}

Thus, the best performance of \textbf{Naive Bayes} was \textbf{90.88\%} but word embedding wasn't compatible and hence wasn't used but it obtained a good accuracy when compared to an ensemble model such as Random Forest.

\subsubsection{\textbf{\textit{Gradient Boost Analysis}}}
The two feature selection extraction were used along-with Gradient Boosting Classifier. The number of estimators used were 100, and the boosting took place for 100 epochs 
	\begin{table}[!htbp]
\begin{center}
    \begin{tabular}{ |c|c|c| } 
    \hline
     \textbf{Metric (in percentage)} & \textbf{TFIDF} & \textbf{Word Embedding} \\
    \hline
     Accuracy & \cellcolor{blue!25}86.86 & 84.90\\
    \hline
     Precision & \cellcolor{blue!25}90.92 & 85.86\\
    \hline
     Recall & 83.21 & \cellcolor{blue!25}85.17\\
    \hline
     F1-Score & \cellcolor{blue!25}86.89 & 85.52\\
    \hline
    \end{tabular}
    \end{center}
        \caption{\label{tab:table-name}Accuracy of Gradient Boost using TFIDF and Word Embedding}
\end{table}

As seen in table IV, the best performance of \textbf{Gradient Boost} using TFIDF to obtain vectors was 86.86\% and with word embedding it was observed as 84.90\%. Hence TFIDF provided better performance with \textbf{86.86\%} accuracy but word embeddings provided higher recall indicating lower false negatives.

\subsubsection{\textbf{\textit{K Nearest Neighbours Analysis}}}
The two feature extraction techniques were used along-with K Nearest Neighbours. The number of neighbours used for this analysis was 6 for both TFIDF and word embedding. 

	\begin{table}[!htbp]
\begin{center}
    \begin{tabular}{ |c|c|c| } 
    \hline
     \textbf{Metric (in percentage)} & \textbf{TFIDF} & \textbf{Word Embedding} \\
    \hline
     Accuracy & \cellcolor{blue!25}90.32 & 86.54\\
    \hline
     Precision & \cellcolor{blue!25}93.02 & 84.89\\
    \hline
     Recall & 88.12 & \cellcolor{blue!25}90.35\\
    \hline
     F1-Score & \cellcolor{blue!25}90.50 & 87.54\\
    \hline
    \end{tabular}
    \end{center}
        \caption{\label{tab:table-name}Accuracy of KNN using TFIDF and Word Embedding}
\end{table}

As seen in the given table, the best performance of \textbf{KNN} using TFIDF to obtain vectors was 90.32\% and with word embedding it was observed as 86.54\%. Hence TFIDF provided better performance with \textbf{90.32\%} accuracy but word embeddings provided higher recall indicating lower false negatives.
\subsection{\textbf{\textit{Deep Learning Models}}}
The standalone models used here were each trained for 10 epochs wheras each model in RMDL was trained for 8 epochs. Training required substantial time ranging from 2 minutes for DNN to 30 hours for RNN.

\subsubsection{\textbf{\textit{Dense Neural Network Analysis}}}
DNNs can be modeled in a variety of ways by introducing different layers of dropout, convolutional and dense layers, in summary we used 4 hidden dense layers with dropout after each layer and this was used to train 9.4M parameters with TFIDF and 3.2M parameters with Glove word embeddings after an addition of flattening layer before the output layer.   

	\begin{table}[!htbp]
\begin{center}
    \begin{tabular}{ |c|c|c| } 
    \hline
     \textbf{Metric (in percentage)} & \textbf{TFIDF} & \textbf{Word Embedding} \\
    \hline
     Accuracy & \cellcolor{blue!25}93.87 & 92.47\\
    \hline
     Precision & 94.03 & \cellcolor{blue!25}95.10\\
    \hline
     Recall & \cellcolor{blue!25}94.28 & 90.26\\
    \hline
     F1-Score & \cellcolor{blue!25}94.15 & 92.62\\
    \hline
    \end{tabular}
    \end{center}
        \caption{\label{tab:table-name}Accuracy of DNN using TFIDF and Word Embedding}
\end{table}

As seen in table VI, the best performance of \textbf{DNN} using TFIDF to obtain vectors was 93.87\% and with word embedding it was observed as 92.47\%. Hence TFIDF provided better performance with \textbf{93.87\%} accuracy but word embedding provided higher precision indicating higher true positives. 

\subsubsection{\textbf{\textit{Convolutional Neural Network Analysis}}}
Similar to DNN, CNN hosts a variety of models that can be used for our task, we used 6 convolutional layers with varying filter sizes along with a single concatenation layer. Also 5 average pooling layers were used was used to extract important features. For TFIDF 6.5M parameters were used to train compared to 2.2M used with word embedding. 
	\begin{table}[!htbp]
\begin{center}
    \begin{tabular}{ |c|c|c| } 
    \hline
     \textbf{Metric (in percentage)} & \textbf{TFIDF} & \textbf{Word Embedding} \\
    \hline
     Accuracy & 89.11 & \cellcolor{blue!25}93.92\\
    \hline
     Precision & 90.42 & \cellcolor{blue!25}93.80\\
    \hline
     Recall & 88.57 & \cellcolor{blue!25}94.64\\
    \hline
     F1-Score & 89.49 & \cellcolor{blue!25}94.22\\
    \hline
    \end{tabular}
    \end{center}
        \caption{\label{tab:table-name}Accuracy of DNN using TFIDF and Word Embedding}
\end{table}

We obtained an accuracy of 89.11\% using TFIDF by training 6.5M parameters and 93.92\% using word embeddings by training 2.2M parameters. The case we see with CNN is quite different from all the models we analyzed, the performance with word embeddings was much better compared to TFIDF, this can be due to a few reasons. Firstly the input layer used with word embeddings was an Embedding Layer which provided advantage in terms of reducing sparse vectors being created (as a by product of TFIDF) and maintaining word relationships intact during feature selection by convolutional layer. Another factor weighs in is that the dimensions are fixed for TFIDF whereas we can define the dimensions for word embedding which (in this case) proved to perform better. Hence the best accuracy of the two was \textbf{93.92\%}

\subsubsection{\textbf{\textit{Recurrent Neural Network}}}
GRU and LSTM are two generally used Recurrent units in RNN. To gain an insight about the performance for our use case both were deployed having similar model structure with just the recurrent layers changing between the two where for GRU, all recurrent layers were GRU and for LSTM, the recurrent layers were LSTM. Training of both models was also computationally expensive with taking almost \textbf{30 hours} to train a the model in both cases. But here we observe that the performance of both models varies by a hairline for both feature extraction techniques. One reason that there is not much difference is that the dataset used isn't huge and hence doesn't have much vocabulary from which can long sequences can be created and put to use by LSTM. On the other hand lack of data size doesn't use the advantages of GRU having no memory cells. Hence, both perform similarly, but the best performance comes out to be \textbf{92.75\%} for RNN with GRU and TFIDF. We can also notice that the precision for both approaches was better for word embeddings which is as previously seen in most of the other models.
\begin{itemize}
\item \textbf{Gated Recurrent Units:} The structure of RNN comprised of 4 GRU layers each followed by dropout layers and eventually the dense layer to train 1.9M parameters. TFIDF and Word Embeddings were used to extract feature vectors and achieve an accuracy of \textbf{92.75\%} and \textbf{92.14\%}.

\begin{table}[!htbp]
\begin{center}
    \begin{tabular}{ |c|c|c| } 
    \hline
     \textbf{Metric (in percentage)} & \textbf{TFIDF}& \textbf{Word Embedding} \\
    \hline
     Accuracy & \cellcolor{blue!25}92.75 & 92.14 \\
    \hline
     Precision & 92.66 & \cellcolor{blue!25}94.15 \\
    \hline
     Recall & \cellcolor{blue!25}93.67 & 90.62 \\
    \hline
     F1-Score & \cellcolor{blue!25}93.11 & 92.35 \\
    \hline
    \end{tabular}
    \end{center}
        \caption{\label{tab:table-name}Accuracy of RNN using GRU}
\end{table}

\item \textbf{Long Short Term Memory:} The structure of RNN comprised of 4 LSTM layers each followed by dropout layers and eventually the dense layer to train 2.4M. TFIDF and Word Embeddings were used to extract feature vectors and achieved an accuracy of \textbf{92.71\%} and \textbf{92.24\%} respectively.

\begin{table}[!htbp]
\begin{center}
    \begin{tabular}{ |c|c|c| } 
    \hline
     \textbf{Metric (in percentage)} & \textbf{TFIDF} & \textbf{Word Embedding} \\
    \hline
     Accuracy & \cellcolor{blue!25}92.71 &92.24 \\
    \hline
     Precision & 92.57 & \cellcolor{blue!25}93.12 \\
    \hline
     Recall & \cellcolor{blue!25}93.57 & 91.96 \\
    \hline
     F1-Score & \cellcolor{blue!25}93.07 & 92.54 \\
    \hline
    \end{tabular}
    \end{center}
        \caption{\label{tab:table-name}Accuracy of RNN using LSTM}
\end{table}

\end{itemize}
\subsubsection{\textbf{\textit{Random Multimodel Deep Learning }}}
RMDL uses CNNs, DNNs and RNNs, with varying layers, nodes, epochs to introduce normalization in the results leveraging ensemble learning \cite{b25}. It uses TFIDF for training DNN but Glove word embeddings for training CNN and RNN. We set the numbers of models used to 3 each and tuned minimum nodes,  minimum number of layers and dropout to be passed as parameters to the model.

	\begin{table}[!htbp]
\begin{center}
    \begin{tabular}{ |c|c|c| } 
    \hline
     \textbf{Model} & \textbf{Feature Extraction} & \textbf{Accuracy} \\
    \hline
     DNN-0 & TFIDF & 93.50\\
    \hline
     DNN-1 & TFIDF & 93.83\\
    \hline
     DNN-2 & TFIDF & 93.83\\
    \hline
     CNN-0 & Glove Word Embeddings (50D) & 73.27\\
    \hline
     CNN-1 & Glove Word Embeddings (50D) & 73.22\\
    \hline
     CNN-2 & Glove Word Embeddings (50D) & 92.89\\
    \hline
     RNN-0 & Glove Word Embeddings (50D) & 77.28\\
    \hline
     RNN-1 & Glove Word Embeddings (50D) & 91.40\\
    \hline
     RNN-2 & Glove Word Embeddings (50D) & 91.26\\
    \hline
    \end{tabular}
    \end{center}
        \caption{\label{tab:table-name}Accuracy using RMDL}
\end{table}
The above tables provide insights into the accuracy models used where in DNN-x (x is just a index number used to name model). Although RMDL shouldn't be evaluated on a model by model basis, but we see that the performance of models vary greatly in terms of performance which underscores the fact that the hyper parameters ultimately impact the performance of the models in a huge way. The combined accuracy achieved by RMDL was \textbf{92.75\%}
\subsection{\textit{Comparative Study}}
Analysing the best accuracy of each algorithms as shown in Table XI, the best accuracy of \textbf{93.92\%} for classifying tweets in COVID-19 dataset. The major pattern we see that with most models, performance with TFIDF excels with that of Word Embedding. There can be some possible reasons for this observation.
\begin{itemize}
    \item Word Embeddings doesn't have the ability to form relationships between new occurring words and use them for training. Though FastText have the ability to use Bag of Words to form vectors for new vocabulary \cite{b26} . Most used words in the COVID-19 dataset were \textit{COVID-19} and \textit{coronavirus} which prior to the pandemic weren't mainstream except medical articles and papers, this unavailability of corpus containing the above words translated to absence of their vectors and relationships in Glove Word Embeddings. The same doesn't happen for TFIDF as it uses the whole available vocabulary in train data to form the vectors. 
    \item Overfitting is a common scenarios while using word embeddings. As word embedding is a complex form of word representation (in addition to the limited vocabulary) it is very much likely that the train data is overfitted in our case.
    \item Other con of using complex word representation is that they carry more hidden information which is particularly not useful for our case but we see in results indicating that word embeddings employ relationship between words to obtain better precision.
\end{itemize}
We also see a stark difference in deep learning models and conventional models being used, signalling to the fact that deep learning approaches, although computationally expensive can provide greater performance in text classification task.
	\begin{table}[!htbp]
\begin{center}
    \begin{tabular}{ c|c|c } 
    \hline
     \textbf{Model} & \textbf{TFIDF} & \textbf{Word Embedding}  \\
    \hline
     Random Forest & 91.30 & 87.05\\
    \hline
     Naive Bayes & 90.88 & -\\
    \hline
     Gradient Boost & 86.86 & 84.90\\
    \hline
     KNN & 90.32 & 86.54\\
    \hline
     DNN & \cellcolor{blue!25}\textbf{93.87} & 92.47\\
    \hline
     CNN & 89.11 & \cellcolor{blue!25}\textbf{93.92}\\
    \hline
     RNN (GRU) & 92.75 & 92.14\\
    \hline
     RNN (LSTM) & 92.71 & 92.24\\
    \hline
     RMDL & \multicolumn{2}{c}{92.75} \\
    \hline
    \end{tabular}
    \end{center}
        \caption{\label{tab:table-name}Comparison of performance for models used}
\end{table}

\section{Future Work}
This work could be further extended as :
\begin{itemize}
    \item Since our current dataset contains only 10700 samples, to analyze the effectiveness and accuracy of the algorithms, this research could be extended on a larger dataset having larger set of vocabulary.
    \item This project is worked upon by using fake news tweets and tweets of official handles of authorities, it could further be extended using the records of more diverse social media platforms or media outlets to enable a holistic approach to the problem
    \item Word Embeddings used can be from a corpus which contains COVID-19 related terminology to even out any missing vocabulary scenarios.
    \item 2D or 3D Convolutional layers can be used to extract features more efficiently. Further the deep models can be made wider instead of deeper to enhance performance.
\end{itemize}
\section{Conclusion}\hfill
This paper attempts to detect and classify fake news regarding \textit{COVID-19} circulating with rapid pace on social media platform twitter using conventional models and deep learning models with different feature extraction techniques TFIDF and GloVe word Embeddings. We found that the best performance of 93.92\% was achieved by CNN along with word embeddings. Surprisingly, we observed that performance of TFIDF was better than a more complex word representation technique such as word embeddings for almost all the models used except CNN. Overall deep learning models (CNN,RNN, DNN and RMDL) performed better than conventional classification approaches. The outcomes of the analysis will aid institutions and researchers to enhance detecting fake news regarding COVID-19 and tackle it at earliest. We were able to detect fake tweets with the best accuracy of \textbf{93.92\%} using CNN along with Glove word embeddings. This 


\begin{thebibliography}{00}
\bibitem{b1} Nyhan, B., \& Reifler, J. (2010). When corrections fail: The persistence of political misperceptions. Political Behavior, 32(2), 303-330. 

\bibitem{b2} J.Cement. 2020.   Number of social media users 2025.Statista. Accessed: 2021-02-22.

\bibitem{b3} Stefanie  Panke.  2020. Social  media  and  fake  news. aace.

\bibitem{b4} Nasser Karimi and Jon Gambrell. 2020.  Hundreds die of poisoning in iran as fake news suggests methanolcure for virus.Times of Israel. Accessed: 2021-03-31. 

\bibitem{b5} Bernard Marr. Coronavirus Fake News: How Facebook, Twitter, And Instagram Are Tackling The Problem. Forbes. Accessed: 2021-03-22.

\bibitem{b6}  Anumeha Chaturvedi. Covid fallout: WhatsApp changes limit on forwarded messages, users can send only 1 chat at a time. Economic Times India. Accessed: 2021-03-22.

\bibitem{b7} Bridgman, Aengus, et al. "The causes and consequences of COVID-19 misperceptions: Understanding the role of news and social media." Harvard Kennedy School Misinformation Review 1.3 (2020).

\bibitem{b8} Pulido, Cristina M., et al. "COVID-19 infodemic: More retweets for science-based information on coronavirus than for false information." International Sociology 35.4 (2020): 377-392.

\bibitem{b9} Dixon, Graham, and Christopher Clarke. "The effect of falsely balanced reporting of the autism–vaccine controversy on vaccine safety perceptions and behavioral intentions." Health education research 28.2 (2013): 352-359.

\bibitem{b10} WHO warns of Covid-19 fake news; clarifies never predicted deaths in India. Business Today. Accessed: 2021-04-09.

\bibitem{b11} Ramos, Juan. "Using tf-idf to determine word relevance in document queries." Proceedings of the first instructional conference on machine learning. Vol. 242. No. 1. 2003.

\bibitem{b12} Pennington, Jeffrey \& Socher, Richard \& Manning, Christopher. (2014). Glove: Global Vectors for Word Representation. EMNLP. 14. 1532-1543. 10.3115/v1/D14-1162.

\bibitem{b13} Ahmed, Hadeer, Issa Traore, and Sherif Saad. "Detecting opinion spams and fake news using text classification." Security and Privacy 1.1 (2018): e9.

\bibitem{b14} Jamal Abdul Nasir, Osama Subhani Khan, Iraklis Varlamis, Fake news detection: A hybrid CNN-RNN based deep learning approach,International Journal of Information Management Data Insights, Volume 1, Issue 1, 2021, 100007.

\bibitem{b15} Zeng, Jiangfeng \& Zhang, Yin \& Ma, Xiao. (2020). Fake news detection for epidemic emergencies via deep correlations between text and images. Sustainable Cities and Society. 66. 102652. 10.1016/j.scs.2020.102652. 

\bibitem{b16} Granik, Mykhailo and Volodymyr Mesyura. “Fake news detection using naive Bayes classifier.” 2017 IEEE First Ukraine Conference on Electrical and Computer Engineering (UKRCON) (2017): 900-903.

\bibitem{b17} Seo, Yeong-Seok \& Jang, Yonghun \& Park, Chang-Hyeon. (2019). Fake News Analysis Modeling Using Quote Retweet. Electronics. 8. 10.3390/electronics8121377. 

\bibitem{b18} Thota, Aswini; Tilak, Priyanka; Ahluwalia, Simrat; and Lohia, Nibrat (2018) "Fake News Detection: A Deep Learning Approach,"SMU Data Science Review: Vol. 1 : No. 3 , Article 10.

\bibitem{b19} Ibrain, Álvaro \& Lloret, Lara. (2019). Fake news detection using Deep Learning.

\bibitem{b20} Le, Hoa T., Christophe Cerisara, and Alexandre Denis. "Do convolutional networks need to be deep for text classification?." arXiv preprint arXiv:1707.04108 (2017).

\bibitem{b21} Patwa, Parth, \textbf{et al}. "Fighting an infodemic: Covid-19 fake news dataset." arXiv preprint arXiv:2011.03327 (2020).

\bibitem{b22} Webb, Geoffrey I. "Naïve Bayes." Encyclopedia of machine learning 15 (2010): 713-714.

\bibitem{b23} Guo, Gongde, et al. "KNN model-based approach in classification." OTM Confederated International Conferences" On the Move to Meaningful Internet Systems". Springer, Berlin, Heidelberg, 2003.

\bibitem{b24} Yu, Shujuan, et al. "Attention-based LSTM, GRU and CNN for short text classification." Journal of Intelligent \& Fuzzy Systems Preprint (2020): 1-8.
\bibitem{b25} Kamran Kowsari, Mojtaba Heidarysafa, Donald E. Brown, Kiana Jafari Meimandi,\& Laura E. Barnes. 2018. RMDL: Random Multimodel Deep Learning for Classification. Association for Computing Machinery, New York, NY, USA, 19–28.
\bibitem{b26} Bojanowski, Piotr, et al. "Enriching word vectors with subword information." Transactions of the Association for Computational Linguistics 5 (2017): 135-146.
\end{thebibliography}
\end{document}